\documentclass[letterpaper, 10 pt, conference]{ieeeconf}  

\IEEEoverridecommandlockouts                              

\makeatletter
\DeclareRobustCommand\onedot{\futurelet\@let@token\@onedot}
\def\@onedot{\ifx\@let@token.\else.\null\fi\xspace}

\def\ie{\emph{i.e}\onedot} 

\usepackage{graphics} 
\usepackage{balance}
\usepackage{subcaption}
\usepackage{amsmath} 
\usepackage{amssymb}  

\usepackage[style=ieee]{biblatex}
\usepackage{xspace}
\usepackage{xcolor,csquotes}
\makeatletter
\let\NAT@parse\undefined
\makeatother
\usepackage[pdftex,pdfpagelabels,bookmarks,hyperindex,hyperfigures]{hyperref}
\hypersetup{colorlinks,linkcolor={red!50!black},citecolor={blue!50!black},urlcolor={blue!80!black}}
\usepackage[capitalise]{cleveref}

\newtheorem{problem}{Problem}
\DeclareMathOperator*{\argmin}{arg\,min}
\newcommand{\mT}{\mathcal{T}}
\newcommand{\mU}{\mathcal{U}}
\newcommand{\targetVel}{\dot{\sigma}^t}
\newcommand{\copAdmit}{u^d}
\newcommand{\refState}{x^*}
\newcommand{\refCop}{u^*}
\newcommand{\optVel}{\dot{\sigma}^{opt}}
\newcommand{\optCop}{u^{opt}}
\newcommand{\optTime}{T^{opt}}
\newcommand{\targetTime}{T^t}
\newcommand{\refDCM}{\xi^*}
\newcommand{\refSigma}{\sigma^*}
\newcommand{\refVel}{\dot{\sigma}^*}
\newcommand{\refContVar}{reference control variable\xspace}
\newcommand{\userDesVel}{user desired velocity\xspace}
\newcommand{\targetVelocity}{target velocity\xspace}
\newcommand{\qsw}{q^{sw}}
\newcommand{\qsp}{q^{sp}}
\newcommand{\dqsw}{\dot{q}^{sw}}
\newcommand{\dqsp}{\dot{q}^{sp}}
\newcommand{\uniLatContConst}{unilateral contact constraints\xspace}

\usepackage{tikz}
\usetikzlibrary{arrows,arrows.meta}
\usetikzlibrary{calc,positioning,decorations,decorations.pathmorphing,decorations.pathreplacing}
\usetikzlibrary{fit}

\tikzstyle{block} = [draw,rectangle,thick,minimum height=2em,minimum width=2em]
\tikzstyle{connector} = [->,thick]
\tikzstyle{merge} = [draw, circle, thick,minimum height=2em,minimum width=2em]

\title{\LARGE \bf
Enabling safe walking rehabilitation on the exoskeleton Atalante: experimental results
}
\author{Maxime Brunet, Marine Pétriaux, Florent Di Meglio and Nicolas Petit \thanks{M. Brunet and M. Pétriaux are with Wandercraft, 88 Rue de Rivoli, 75004 Paris, France {\tt\small maxime.brunet@wandercraft.eu}}
\thanks{F. Di Meglio and N. Petit are with MINES Paris, Centre Automatique et Systèmes, PSL University, 60 bd. St Michel, 75272 Paris Cedex, France}}

\begin{document}

\maketitle
\thispagestyle{empty}
\pagestyle{empty}

\begin{abstract}
This paper exposes a control architecture enabling rehabilitation of walking impaired patients with the lower-limb exoskeleton Atalante. Atalante’s control system is modified to allow the patient to contribute to the walking motion through their efforts. Only the swing leg degree of freedom along the nominal path is relaxed. An online trajectory optimization checks that the muscle forces do not jeopardize stability. The optimization generates reference trajectories that satisfy several key constraints from the current point to the end of the step. One of the constraints requires that the center or pressure remains inside the support polygon, which ensures that the support leg subsystem successfully tracks the reference trajectory. As a result of the presented works, the robot provides a non-zero force in the direction of motion only when required, helping the patient go fast enough to maintain balance (or preventing him from going too fast).
Experimental results are reported. They illustrate that variations of $\pm 50\%$ of the duration of the step can be achieved in response to the patient's efforts and that many steps are achieved without falling. A video of the experiments can be viewed at \url{https://youtu.be/_1A-2nLy5ZE}.

\end{abstract}

\section{Introduction}

Patients suffering from walking impairments are unable to produce the efforts required to achieve regular walk patterns. High-dosage walking rehabilitation has many benefits but is laborious for physiotherapists, particularly for the most impaired patients who struggle maintaining their balance. From these observations, the concept of robotic-assisted gait training has emerged. This concept involves motorized devices (in this paper a robotic exoskeleton) and aims at teaching patients how to produce the appropriate efforts to walk, following either a nominal gait pattern, ideally, or custom gait patterns tailored to their specific disabilities. All robotic-rehabilitation control laws~\cite{artemiadisReviewRobotAssistedLowerLimb2020,liReviewControlStrategies2021,martinezControllerGuidingLeg2018,martinezVelocityFieldBasedControllerAssisting2019} introduce a certain level of freedom given to the patients. For exoskeletons, this constitutes a challenge because the patient muscle forces may jeopardize walk stability in unpredictable ways.

In this paper, we present control system updates enabling the use of the self-balanced lower-limb exoskeleton Atalante for rehabilitation. We choose to let the patient physically contribute to the motion of the swing leg and, consequently, we allow a modulation of the velocity at which a predefined gait is traveled. Because this directly impacts stability, a new reference trajectory has to be computed for the low-level controllers of the whole system. This is done online by an optimization-based trajectory planning algorithm. 
This methodology is presented in the article and tested experimentally. The two legs of Atalante are the subjects of very distinct changes.

On the swing leg we proceed as follows. The nominal gait for the two-legged system, generated as in~\cite{haribFeedbackControlExoskeleton2018}, serves to define a geometric path for the swing leg. The exoskeleton efforts in the longitudinal direction of the path are nullified and left for the patient to produce, while the robot motion is strictly controlled in the hyperplane orthogonal to the path. To this end, we rely on the Virtual Guides~(VG) methodology~\cite{jolyImposingMotionConstraints1995,sanchezrestrepoIntuitiveIterativeAssisted2018}. Resembling~\cite{shuModulationProstheticAnkle2022}, our VG approach maps the high-dimensional user efforts to a one-dimensional quantity: the velocity at which the swing leg's geometric path is followed. Implicitly, this defines a new schedule for the path, the \emph{patient schedule}, which we aim to follow as long as safety is not threatened.  

On the support leg the control structure is also changed. The Atalante control system uses admittance~\cite{caronStairClimbingStabilization2019} to generate the contact forces that ensure the stabilization of the Center-of-Mass (CoM) dynamics around a reference trajectory. The reference trajectory tracked must satisfy the unilateral contact constraints (as a consequence, the Center of Pressure (CoP) must remain in the support polygon). Because the swing leg degrees of freedom (DoF) are used for the rehabilitation task, only the support leg can be used for the admittance controller.

To define this reference trajectory consistently with the patient schedule, a simple time rescaling of the nominal gait is possible but may violate of the aforementioned constraints. Therefore, combining VG and admittance control entails defining a more careful approach. We chose to adopt an {online planning strategy}.
This approach is similar in spirit to~\cite{tuckerPreferenceBasedLearningExoskeleton2020}, as we optimize the trajectory to best satisfy the patient's input, with the difference that we adapt the trajectory at a higher frequency, to maximize responsiveness to the patient efforts

The online planning strategy we implement solves an optimal control problem (OCP) over an unspecified horizon for a Linear Inverted Pendulum (LIP) model, which represents the overall balance dynamics of the system. The patient schedule is treated as a penalty on the final time.
The \uniLatContConst is a constraint on the CoP.
To ensure the feasibility of the next step, it is sufficient (see~\cite{wieberModelingControlLegged2016}) to require that the trajectory endpoint satisfies some geometric constraints. These are included in the OCP. 
Similar online trajectory generation can be found in many related works 
for quadrupeds~\cite{bellicosoDynamicLocomotionOnline2018,jeneltenTAMOLSTerrainAwareMotion2022,grandiaPerceptiveLocomotionNonlinear2022,jeonOnlineOptimalLanding2022,bjelonicOfflineMotionLibraries2022,mastalliAgileManeuversLegged2022, mastalliFeasibilitydrivenApproachControllimited2022}, humanoids \cite{dantecWholeBodyModel2021,romualdiOnlineNonlinearCentroidal2022a}, or manipulators~\cite{kleffHighFrequencyNonlinearModel2021}. It is worth noting that
of all these schemes, only~\cite{kleffHighFrequencyNonlinearModel2021,mastalliCrocoddylEfficientVersatile2020} solves a non-linear MPC problem at 1kHz as we do (with a CPU twice as fast as Atalante onboard computer). The simplicity of the LIP model allows us to use a fast resolution method for this nonlinear OCP, relying on the theoretical study of~\cite{brunetFastReplanningLowerlimb2022a}. 

The methodology has been experimentally tested. During the experiments, the optimization algorithm finds a CoM trajectory. As long as the duration of this trajectory matches the patient schedule, the user is allowed to drive the swing leg velocity. Otherwise, the patient schedule is considered non-feasible (\ie it jeopardizes the balance of the system). It is overridden and the swing leg velocity is modified. Consequently, the swing leg actuators  generate non-zero forces in the direction of motion, helping the patient go fast enough to maintain balance (or preventing him from going too fast). In all cases, the support leg controls the robot balance. As a result of the methodology, the robot assists the swing leg motion only when required.

The main contribution of the article is the detailed exposition of this control methodology, along with its experimental validation. In particular, we report an experiment consisting of a succession of 10 steps with a user-driven velocity variation of more than $50\%$ of the baseline velocity.

The paper is organized as follows. In~\cref{sec:controllers} we describe the dynamical models used for control design, the swing leg Virtual Guides controller,  and the support leg admittance controller. In~\cref{sec:highlevelctrl}, we describe the online planning strategy generating the controllers' reference trajectories and compare it with a naive approach. In~\cref{sec:experiments} we perform experiments to illustrate quantitatively the safety increase provided by the online planning strategy, and  the performance of the overall approach.

\section{Feedback controllers: a split-leg design}\label{sec:controllers}
In this section, we describe the Virtual Guide controller used to regulate the swing leg DoF and the admittance controller regulating the contact forces through the support leg actuators. First, we recall the equations of motion on which we rely throughout the article. 
\subsection{Dynamics of the patient-exoskeleton system}\label{sec:models}
\begin{figure}[tb]
     \begin{subfigure}[b]{0.235\textwidth}
        \centering
        \includegraphics[width=3.5cm]{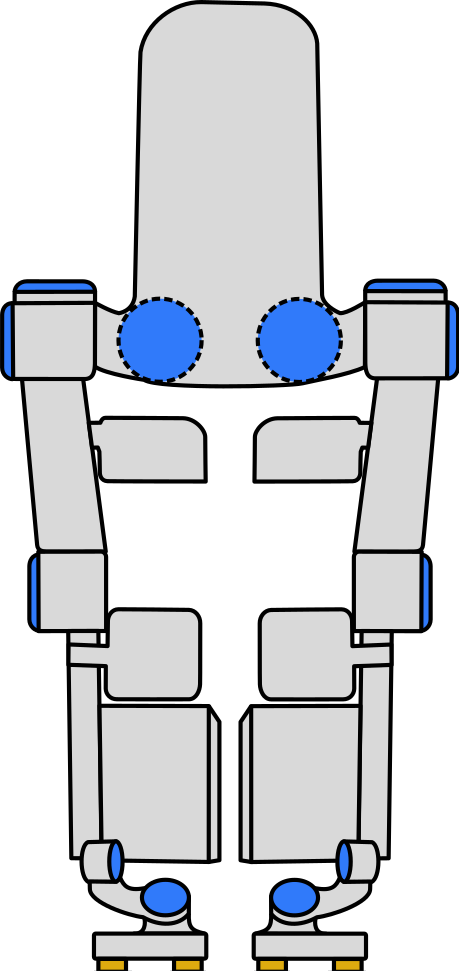}
    \end{subfigure}
     \begin{subfigure}[b]{0.235\textwidth}
        \centering
        \includegraphics[width=3cm]{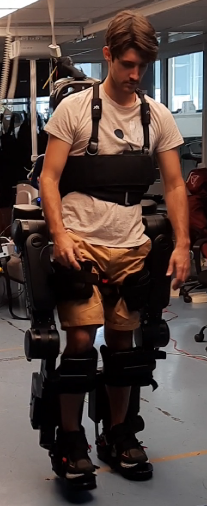}
    \end{subfigure}
    \caption{Atalante's kinematics (revolute joints in blue)-left. An able-bodied user performs a rehabilitation exercise-right.}
    \label{fig:atalanteScheme}
\end{figure}
Consider the exoskeleton depicted in~\cref{fig:atalanteScheme}. The fastening system of the exoskeleton completely assigns the positions of the lower limbs of the patient with respect to the robot. Their torso is less firmly attached, but we consider it rigidly fixed to the exoskeleton. These positions define the major part of the patient's weight distribution, hence we neglect all the other DoF of the patient. However, the mass distribution of patients cannot be easily measured. As a surrogate, we assume they follow a normal distribution, which can be found in~\cite{winterBiomechanicsMotorControl2009}.
Under these assumptions, the patient-exoskeleton system can be modeled as an articulated rigid-body system of total mass~$m$ with 12 actuated DoF (the joints of the exoskeleton) and 6 unactuated DoF (the position and orientation of the exoskeleton pelvis in the world frame). In details, 
the Lagrangian dynamics of the system write~\cite{wieberModelingControlLegged2016}
\begin{equation}\label{lagrangianDyn}
    M(q)\ddot{q} + C(q,\dot{q}) = \begin{pmatrix}
        \tau^{sw} + \tau_u^{sw}\\
        \tau^{sp} + \tau_u^{sp}\\
        0
    \end{pmatrix} + \sum_i J_i(q)^\top f_i
\end{equation}
with $q = \begin{pmatrix}
    \qsw,
    \qsp,
    q_{un}
\end{pmatrix}^\top\in \mathbb{R}^{18}$ the generalized coordinate vector, composed of the actuated positions vector $\qsw$ of the swing leg, $\qsp$ the actuated positions of the support leg, and the unactuated degrees of freedom $q_{un}$, $M(q)$ the generalized inertia matrix of the system, $C(q,\dot{q})$ the combined gravity and inertia effects vector, $(\tau^{sw}, \tau^{sp})$ the vectors of exoskeleton swing and support joint torques to be chosen by the controllers, $(\tau_u^{sw},\tau_u^{sp})$ the vector of  swing and support joint torques created by the patient, $f_i\in\mathbf{R}^3$ the external forces and  $J_i(q)$ the associated Jacobian matrices at each contact point $p_i$.

The dynamics of the system, in an inertial reference frame, taken as a whole, give the Newton and Euler equations~\cite{wieberModelingControlLegged2016}
\begin{equation}\label{newtonEuler}
    \begin{gathered}
    m(\ddot{c} + g) = \sum_i f_i, \quad
    \dot{L} = \sum_i (p_i-c)\times f_i
    \end{gathered}
\end{equation}
with $c = \begin{pmatrix}
    c^{x,y}, c^z
\end{pmatrix}\in\mathbf{R}^3$ the CoM of the system and $L$ the angular momentum of the system with respect to its CoM.

\subsection{Virtual guides controller on the swing leg}\label{sec:virtualGuides}
The methodology of Virtual Guides~\cite{jolyImposingMotionConstraints1995} allows a parametric curve $P$ to be followed at a velocity prescribed by the patient's efforts on the robot. We propose to use it to estimate the user intent.
\subsubsection{Constructing the parametric curve} The parametric curve used by the controller is built from a nominal gait trajectory $\mT : t \in [0, T_f] \mapsto \mT(t) \in \mathbb{R}^{12}$, readily computed as in~\cite{gurrietRestoringLocomotionParaplegics2018}, from which we extract the swing leg trajectory $\mT_{sw} : t\mapsto \mT_{sw}(t) \in \mathbb{R}^{6}$. The latter can be reparametrized with respect to its curvilinear abscissa $\text{s} : \tau \mapsto \int_0^\tau ||\dot{\mT_{sw}}(t)||_2 dt \in [0,L_{max}]$, with $L_{max}$ the total length (\emph{in the articular space}) of the swing leg trajectory. By this formula, the curvilinear abscissa $\text{s}$ is monotonous, therefore, assuming further that the Euclidean norm $||\dot{\mT_{sw}}(t)||_2$ is non-zero for all $t$, $\text{s}$ can be inverted. Then, we can define the parametric curve as $$P \triangleq \mT_{sw}\circ \text{s}^{-1}$$
\subsubsection{Virtual guides low-level controller} In the following $$\sigma: t \mapsto \sigma(t) \in [0,L_{max}]$$ is a freely chosen control variable which defines the current set-point $P(\sigma(t))$ for the swing leg.

The Virtual Guides methodology minimizes the interacting force between the robot and the user while constraining the robot to the parametric path ${P}$. Defining the parametric path in joint-space, by constrast with Cartesian space, enforces inter-joint coordination. At the current point $P(\sigma(t))$, the Frenet-Serret unit tangent vector to the curve~$P$, pointing in the direction of motion is$$T(\sigma) \triangleq \frac{d {P}}{d\sigma} (\sigma)$$ To provide contraction property in the direction orthogonal to~$T$, the joint torques of the swing leg are computed using a high-gain proportional-derivative controller
\begin{equation}\label{PD}
    \tau^{sw}(\sigma, \dot{\sigma}) = K_p^{sw}(P(\sigma) - \qsw) + K_d^{sw}(T(\sigma)\dot{\sigma} - \dqsw)
\end{equation}
with $K_p^{sw}, K_d^{sw} \succ 0$ constant gain matrices.

\subsubsection{Estimation of the \targetVelocity} Following the Virtual Guides approach, we define the estimate of the \userDesVel $\targetVel$ such that the projection of the efforts~$\tau^{sw}(\sigma, \targetVel)$ along the path ${P}$ is nullified, which reads
\begin{align}\label{needOfRehab}
    T(\sigma)^\top\tau^{sw}(\sigma, \targetVel) = 0
\end{align}
This yields~\cite{sanchezrestrepoIntuitiveIterativeAssisted2018}
\begin{equation}\label{VG}
    \targetVel \triangleq \frac{T(\sigma)^\top \left[K_p^{sw}(\qsw - P(\sigma)) + K_d^{sw}\dqsw\right]}{T(\sigma)^\top K_d^{sw} T(\sigma)}
\end{equation}
For the rest of the article, we consider that satisfying the user desire by imposing $\dot{\sigma} = \targetVel$ is the rehabilitation objective. We now call $\targetVel$ the \textit{target velocity}. The design of a safe $\sigma$ such that $\dot \sigma$ is as close to $\targetVel$ as possible is the subject in~\cref{sec:highlevelctrl}.

\subsection{Admittance controller on the support leg}\label{admittance}
The only terms yet to be defined in~\cref{lagrangianDyn}  are the torques of the support leg~$\tau^{sp}$, this is addressed below. 

The Newton-Euler equations~\eqref{newtonEuler} can be simplified into the Linear Inverted Pendulum (LIP) dynamics to ease the stability analysis and the design of controllers. 
Indeed, along gait patterns of moderate velocity, the angular momentum variations are small and can be neglected.
Assuming the robot walks on horizontal ground and the CoM remains at a constant height $c^z$, \cref{newtonEuler} simplifies to the following LIP dynamics~\cite{wieberModelingControlLegged2016}
\begin{equation}\label{eq:LIPdyn}
    \ddot{c}^{x,y} = \omega^2(c^{x,y}-u)
\end{equation}
where~$\omega \triangleq \sqrt{\frac{g}{c^z}}$ is the angular-frequency, and the CoP $u \triangleq \frac{\sum p_i^{x,y} f_i^z}{\sum f_i^z}$ lies inside the support polygon (rectangle)~$\mathcal{U}$ by definition. We note $x$ the state of the LIP $$x(t)\triangleq(c(t),\dot{c}(t))$$

\cref{eq:LIPdyn} reproduces the unstable nature of the system. For stabilization, a State-of-the-art admittance controller such as the one detailed in~\cite{caronStairClimbingStabilization2019} is used
\begin{equation}\label{copReference}
    \copAdmit = \refCop - (1 + \frac{k_p}{\omega}) (\refDCM - \xi) - \frac{k_i}{\omega}\int(\refDCM - \xi) + k_d(\dot{\refDCM} - \dot{\xi})
\end{equation}
with $\xi \triangleq c + \frac{\dot{c}}{\omega} \in \mathbb{R}^2$ the Divergent Component of Motion (DCM), readily computed from the LIP state $x$, the DCM reference trajectory $\refDCM$, computed from the state reference trajectory $\refState$, and the associated CoP reference trajectory $\refCop$, and $k_p, k_d, k_i \succ 0$ three diagonal matrices. The choice of the reference and feedforward trajectories $(\refState,\refCop)$ is the topic of~\cref{sec:highlevelctrl}.

For implementation, $\copAdmit\in \mathbb{R}^2$ is converted into articular targets $({\qsw}^*, {\qsp}^*)\in \mathbb{R}^6\times\mathbb{R}^6$ using inverse kinematics and admittance tasks inspired from~\cite{caronStairClimbingStabilization2019}.
Finally, the support leg joint torques $\tau^{sp}$ are computed using a high-gain proportional-derivative controller
\begin{equation}\label{PDsp}
    \tau^{sp} =  K_p^{sp}({\qsp}^* - \qsp) + K_d^{sp}(\dot{q}^{sp^*} - \dqsp)
\end{equation}
with $K_p^{sw}, K_d^{sw} \succ 0$ constant gain matrices.
More details on this admittance scheme can be found in~\cref{LLCAdmittance}.

\subsection{Summary of low-level controller updates}
\begin{figure}[tb]
    \centering
    \resizebox{\linewidth}{!}{
        \begin{tikzpicture}[node distance=1cm and 1.5cm, >=latex']
            \node[block, align=center] (REF) at (0,0)  [minimum height=1.8cm] {Reference trajectory\\ computation};
            \node[block, right=2cm of REF] (PD) [yshift=4ex]{Swing leg ctrl.~\eqref{PD}};
            \node[block, below=0.5cm of PD] (PDsp) {Support leg ctrl.~\eqref{PDsp}};
            \node[block, right=6.5cm of REF] (ROBOT) [minimum height=1.8cm] {Exoskeleton-patient};
            \node[block, below=0.8cm of PDsp] (VG) {Patient schedule estimator~\eqref{VG}};


            \draw [draw,->] (ROBOT.north) |- ([yshift=5ex]PD.east) -| (PD.north) node[pos=0.5, above, xshift=15ex] {$(\qsw,\dqsw)$};
            \draw [draw,->] (REF.east) -- ++ (0.4, 0) |- (PD.west) node[pos=0.5, above, xshift=6ex] {$\refSigma$};
            \draw [draw,->] (REF.east) -- ++ (0.4, 0) |- (PDsp.west) node[pos=0.5, above, xshift=5ex] {$(\refState, \refCop)$};
            \draw [draw,->] (PD.east) -- ([yshift=4ex]ROBOT.west) node[pos=0.7, above] {$\tau^{sw}$};
            \draw [draw,->] (PDsp.east) -- ([yshift=-3.8ex]ROBOT.west) node[pos=0.6, above] {$\tau^{sp}$};
            \draw [draw,->] (ROBOT.south) |- ([yshift=-5ex]PDsp.east) -|  (PDsp.south) node[pos=0.5, above, xshift=14ex] {$(x, u)$};
            \draw [draw,->] (ROBOT.east)-- ++ (0.25, 0) |-  (VG.east) node[pos=0.5, above, xshift=-10ex] {$(\qsw,\dqsw)$};
            \draw [draw,->] (VG.west) -| (REF.south) node[pos=0.25, above] {$\targetVel$};

        \end{tikzpicture}
    }
    \caption{Proposed control architecture for rehabilitation.}
    \label{BlockDiagram}
\end{figure}
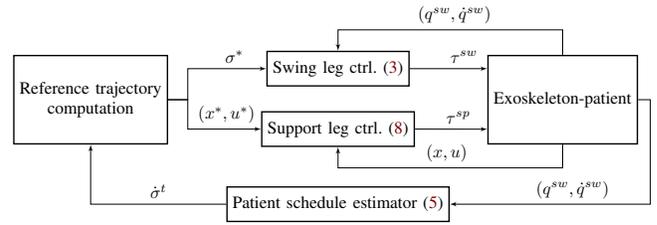

The updates of the control law that we propose are schematically depicted on~\cref{BlockDiagram}. They consist in separated calculations of $\tau^{sw}$ and $\tau^{sp}$, during single support phases. Classically, the admittance methodology is used on both legs during double support phases~(see~\cref{LLCAdmittance} for more details). On the one hand, a new calculation of $\tau^{sw}$ is proposed according to the VG law \cref{PD}. On the other hand, $\tau^{sp}$ is calculated according to the admittance control law \cref{PDsp}. These controllers require a reference trajectory $(x^*,u^*)$, satisfying unilateral contact constraints (the CoP should remain in the support polygon $\mU$), and a schedule~$\sigma^*$, taking into account the patient input~$\targetVel$. Their design is the topic of the next section.
 



\section{Reference trajectory design}\label{sec:highlevelctrl}

In this section, we first expose a Time Rescaling (TR) strategy for the choice of the reference variables $(\refSigma, \refState, \refCop)$. As will appear, it is not sufficient as it does not enforce the unilateral contact constraint.
Then, we expose an Online Planning (OP) strategy which  explicitly takes the constraints into account. We illustrate its benefits on an example. A more thorough experimental investigation of the stability benefits is provided in~\cref{openLoopXP}.

\subsection{Time Rescaling (TR) strategy: a naive approach}\label{nominalCoMTrajLimits}
A natural way to define the \refContVar~$\refSigma$ is to simply integrate the \targetVelocity
\begin{equation}\label{sigmaRescaling}
    \refSigma(t) = \int_0^t \targetVel(\tau) d\tau
\end{equation}
where~$\targetVel$ is given by~\cref{VG}.
Then, from the articular nominal gait $\mT$, and~\cref{sigmaRescaling}, the state reference trajectory $\refState$ can be computed using the time-rescaled nominal gait trajectory and Forward Kinematics (FK), see~\cite{carpentierPinocchioLibraryFast2019}
\begin{equation}
    \refState = FK\circ\mT\circ\text{s}^{-1}\circ\refSigma
\end{equation}
Finally, the corresponding input $\refCop$ can be readily computed from $\refState$ using~\cref{eq:LIPdyn}.
 
However, this TR strategy does not take the \uniLatContConst into account. As a result, the feedforward trajectory $\refCop$ is not confined to the support polygon~$\mathcal{U}$ (and the state reference trajectory $\refState$ does not respect the input-constrained LIP dynamics). 

\begin{figure}[tb]
    \centering
        \includegraphics[scale=0.185]{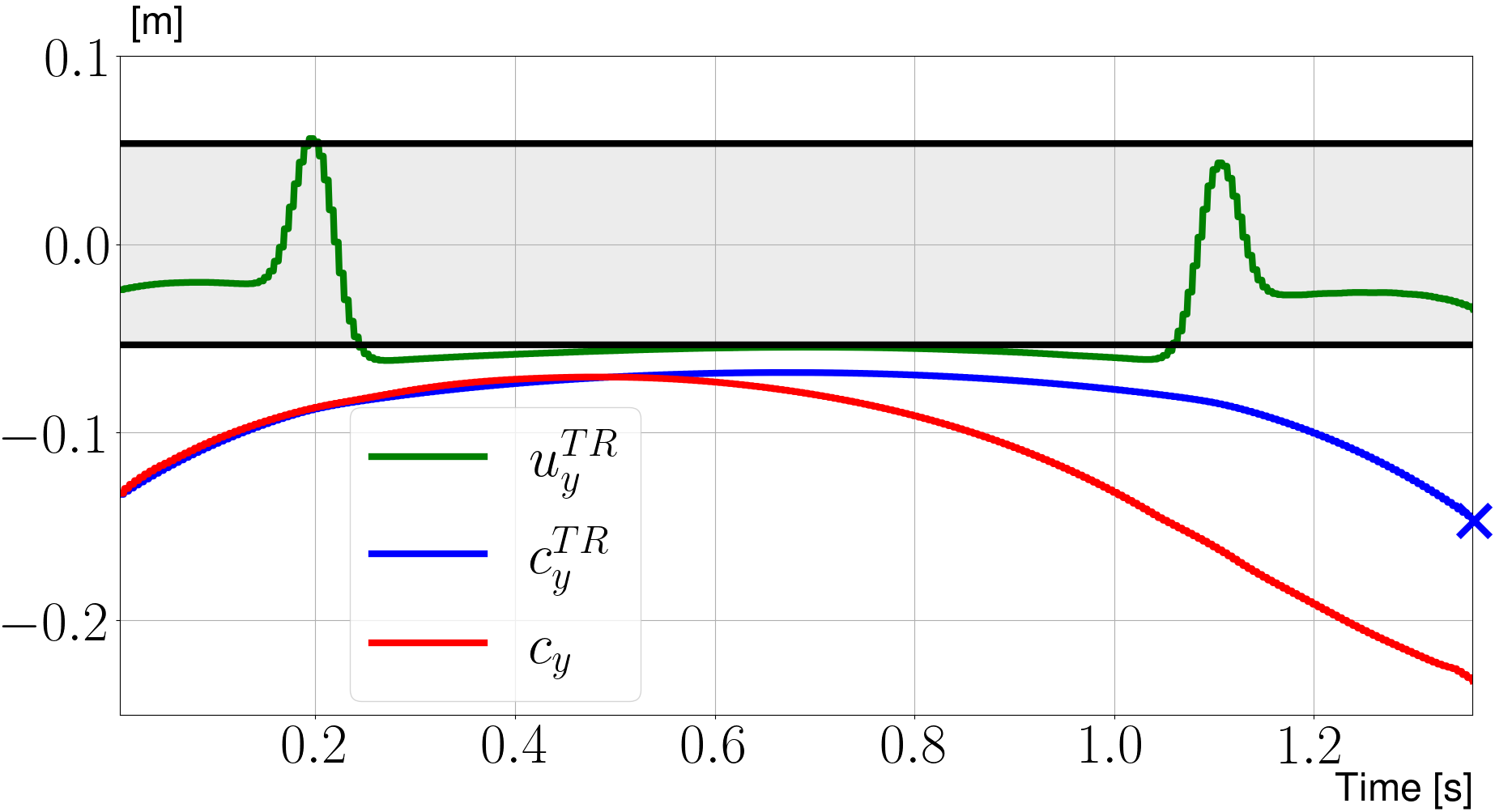}
    \caption{Reference CoP $u^{TR}_y$, CoM $c^{TR}_y$ and measured CoM $c_y$ positions along the Y axis of the inertial frame. Reference quantities computed using the TR strategy and a simulated \targetVelocity as low as 60$\%$. Black horizontal lines represent the support polygon limits.}
    \label{fig:79pc_CoMDCM}
 \end{figure}

This shortcoming is illustrated by a simulation in~\cref{fig:79pc_CoMDCM}. The reference CoP, in green, is not contained in the support polygon when modulating the trajectory at $60\%$ of the nominal velocity. The final state (the endpoint of the red line) obtained by forward integration of the full-state dynamics~\eqref{lagrangianDyn}, using the open-source simulator Jiminy~\cite{duburcqJiminyFastPortable2019}, is different from the nominal final state $x_f \triangleq FK\circ\mT(T_f)$ (the endpoint of the blue line). 
The eight-centimeter resulting error is sufficient to make the robot fall at the end of the step.

\subsection{Online Planning (OP) strategy}
Instead of the previous naive approach, we consider a joint optimization of the variables $\refSigma$ and $(\refState,\refCop)$ taking into account the constraints $\mathcal{U}$ and a next-step LIP-feasibility~\cite{wieberModelingControlLegged2016} constraint $x_f$.

The \targetVelocity $\targetVel$ (saturated to be strictly positive) is converted into a target time $\targetTime$ to the end of the current step
\begin{equation}\label{TFromSigma}
    \targetTime = \frac{L_{max} - \sigma}{\targetVel}
\end{equation}
This equation does not exploit any behavioral description of the patient~\cite{ivanenkoTwothirdsPowerLaw2002}, but solely assumes that the patient's desire is to keep the velocity constant until the end of the step.
A bi-level trajectory optimization problem for the input-constrained LIP dynamics is formulated from the current state $x_0$ as follows
\begin{problem}\label{PB1}
    Given $(x_0, x_f)$ and $\targetTime$ find $\optCop$ and $\optTime$ as
    \begin{equation*}\label{pb}
        \begin{aligned}
            \optTime = &\argmin_{T\in\mathbb{T}(x_0,x_f)} |T-\targetTime|\\
            &\text{s.t.} \ \optCop = \argmin_{u\in\Omega(x_0,x_f,T)}\int_0^T u^2 dt
        \end{aligned}
    \end{equation*}
\end{problem}
where $\mathbb{T}(x_0,x_f)\subset \mathbb{R}$ is the set of times for which $\Omega$ is not empty, $\Omega$ is the set of feasible commands respecting the boundary conditions $(x_0, x_f)$
\begin{equation*}
    \Omega(x_0,x_f,T) \triangleq \left\{u \in U_{ad}(T),\ x^u(0) = x_0,\ x^u(T) = x_f\right\}
\end{equation*}
with $x^u$ the solution of~\cref{eq:LIPdyn} from $x_0$ and $U_{ad}(T) \triangleq \left\{u\  \text{s.t.}\ \forall t\in[0,T],\ u(t) \in \mU\right\}$ is the set of admissible controls.

\Cref{PB1} is in fact a minimal time problem for an input-constrained linear dynamics of dimension~$4$. Its phase plane analysis, decoupling the X and Y directions, has been performed in~\cite{brunetFastReplanningLowerlimb2022a}, covering all possible cases of initial and final conditions. The main finding of~\cite{brunetFastReplanningLowerlimb2022a} is that $\mathbb{T}$ is the union of at most two intervals, such that the solutions $\optTime$ and $\optCop$ can be easily determined using a bisection method, granting high-numerical efficiency. Computation times are reported in table~\cref{computationTimeTable}. They allow a $1$\,kHz update of the reference trajectory. This is consistent with our aim to maximize responsiveness to the patient efforts.
\begin{figure}[tb]
   \begin{center}
      \begin{tabular}{ |c|c|c|c| } 
         \hline
         & min & max & mean \\ 
         \hline
         CPU time & 0.039\,\textrm{ms} & 0.22\,\textrm{ms} & 0.11\,\textrm{ms} \\
         \hline
      \end{tabular}
   \end{center}
   \caption{CPU time of the OP strategy (running on a i7-1185G7E at fixed $1.8$\,\textrm{GHz} frequency).}
   \label{computationTimeTable}
\end{figure}

Then, the \refContVar $\refSigma$ is computed from the optimal time $\optTime$ as follows
\begin{equation*}
    \refSigma = \int_0^t \left(\optVel \triangleq \frac{L_{max} - \sigma(\tau)}{\optTime(\tau)}\right) \ d\tau
\end{equation*}
The reference trajectories $(\refState, \refCop)$ are obtained from the optimal command $\optCop$ as follows
\begin{equation*}
    \refState = x^{\optCop}, \quad
    \refCop = \optCop
\end{equation*}

\begin{figure}[tb]
    \centering
        \includegraphics[scale=0.185]{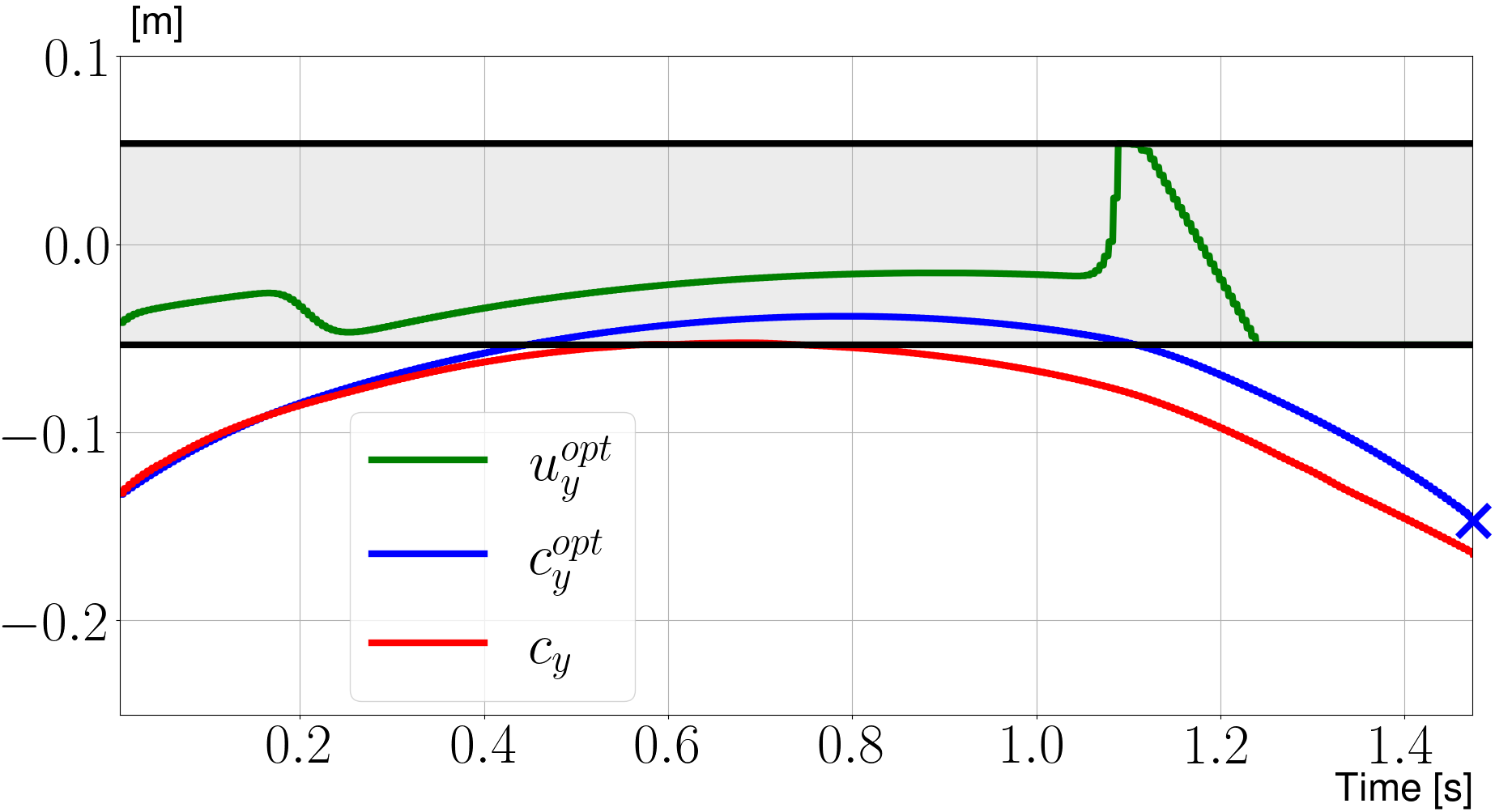}
     \caption{Reference CoP $u^{opt}_y$, CoM $c^{opt}_y$ and measured CoM $c_y$ positions along the Y axis of the inertial frame. Reference quantities computed using the OP strategy and a simulated \targetVelocity as low as 60$\%$. Black horizontal lines represent the support polygon limits.}
    \label{fig:79pc_CoMDCM_safe}
 \end{figure}

The effect of this OP strategy is illustrated on~\cref{fig:79pc_CoMDCM_safe}, where, as opposed to~\cref{fig:79pc_CoMDCM}, the CoP reference trajectory (green) is entirely contained in the support polygon (black horizontal lines) and offset towards the $y=0.0$ line. As a result, the endpoint of the forward integration of the full-state dynamics~\eqref{lagrangianDyn} (the endpoint of the red line) is close to the nominal final state (the blue cross): the state is successfully driven to the final state using the stabilization controlled~\cref{admittance}. This recursively ensures the success of the walk. To further illustrate the merits of this approach, we report experimental results in the next section.

\section{Experimental results}\label{sec:experiments}

Two types of experiments are conducted. First, we compare the stability properties of the two strategies and conclude on the vast superiority of the OP strategy. Then, we explore the performance of the OP strategy in terms of compliance to the patient schedule.

\subsection{Stability comparison}\label{openLoopXP}

We perform several stability comparisons replacing the patient with a dummy.
To simulate the behavior of the patient, we consider pre-recorded\footnote{\ie replacing~\cref{VG} with a predefined function of time} piecewise constant velocity signals~$\targetVel(t)$ consisting of a square wave whose duration and magnitude are varied. An experiment consists of a 10 steps walk in straight line. A practitioner keeps hold of the two lateral exoskeleton handles and is allowed to create an effort with one finger on each hand only. This creates a very low upper-bound on the external forces.
 
\begin{figure}[tb]
    \centering
    \includegraphics[scale=0.26]{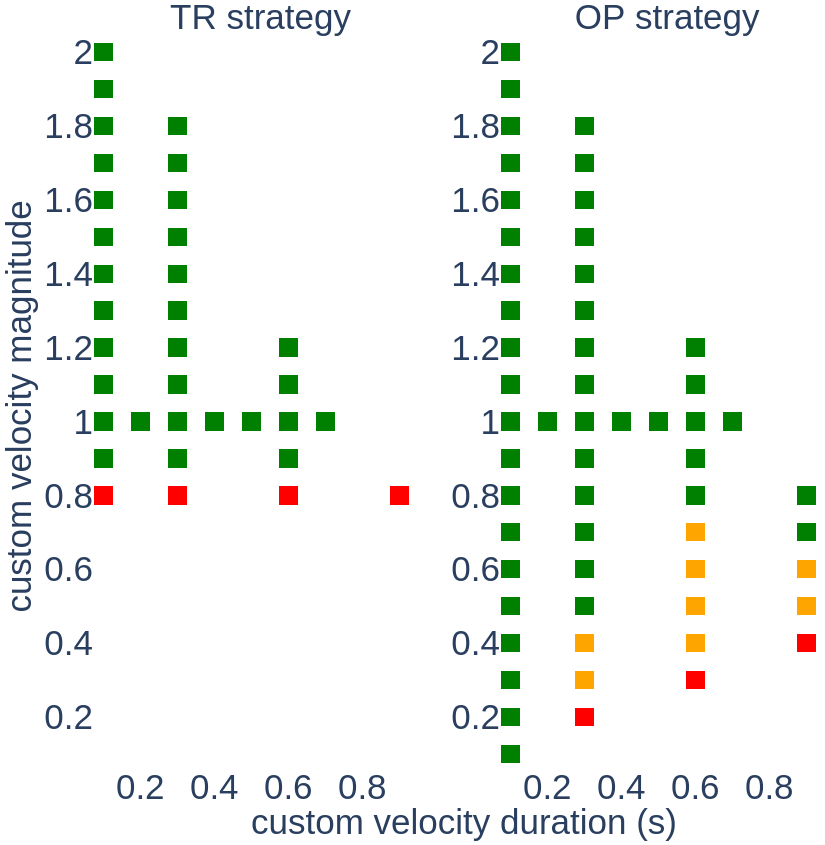}
     \caption{Comparison of experimental stability for velocity variations having various durations and magnitudes\protect\footnotemark. Green: stable without external help. Orange: stable with practitioner help. Red: unstable. Left: TR strategy. Right: OP strategy.}
    \label{fig:quant_results}
 \end{figure}

The reported results on~\cref{fig:quant_results} show a great safety improvement offered by the OP strategy in the low-velocity range, below $90\%$ of nominal velocity, compared to the TR strategy. They stress that, using the OP strategy, the proposed controller is completely preserving the balance of the system for velocities about as low as $70\%$ of the nominal velocity, and velocities as low as $50\%$ of the nominal velocity provided the change duration is lower or equal to $300$\,ms. These results are in complete alignment with the previous stability results obtained in simulation and reported in~\cite{brunetFastReplanningLowerlimb2022a}.
\footnotetext{The white spaces in this figure corresponds to unfeasible values of the parameters violating the constraint $\sigma^*\leq L_{max}$.}

\subsection{Rehabilitation: experiment with an able-bodied user}

\begin{figure}[tb]
        \centering
        \includegraphics[scale=0.18]{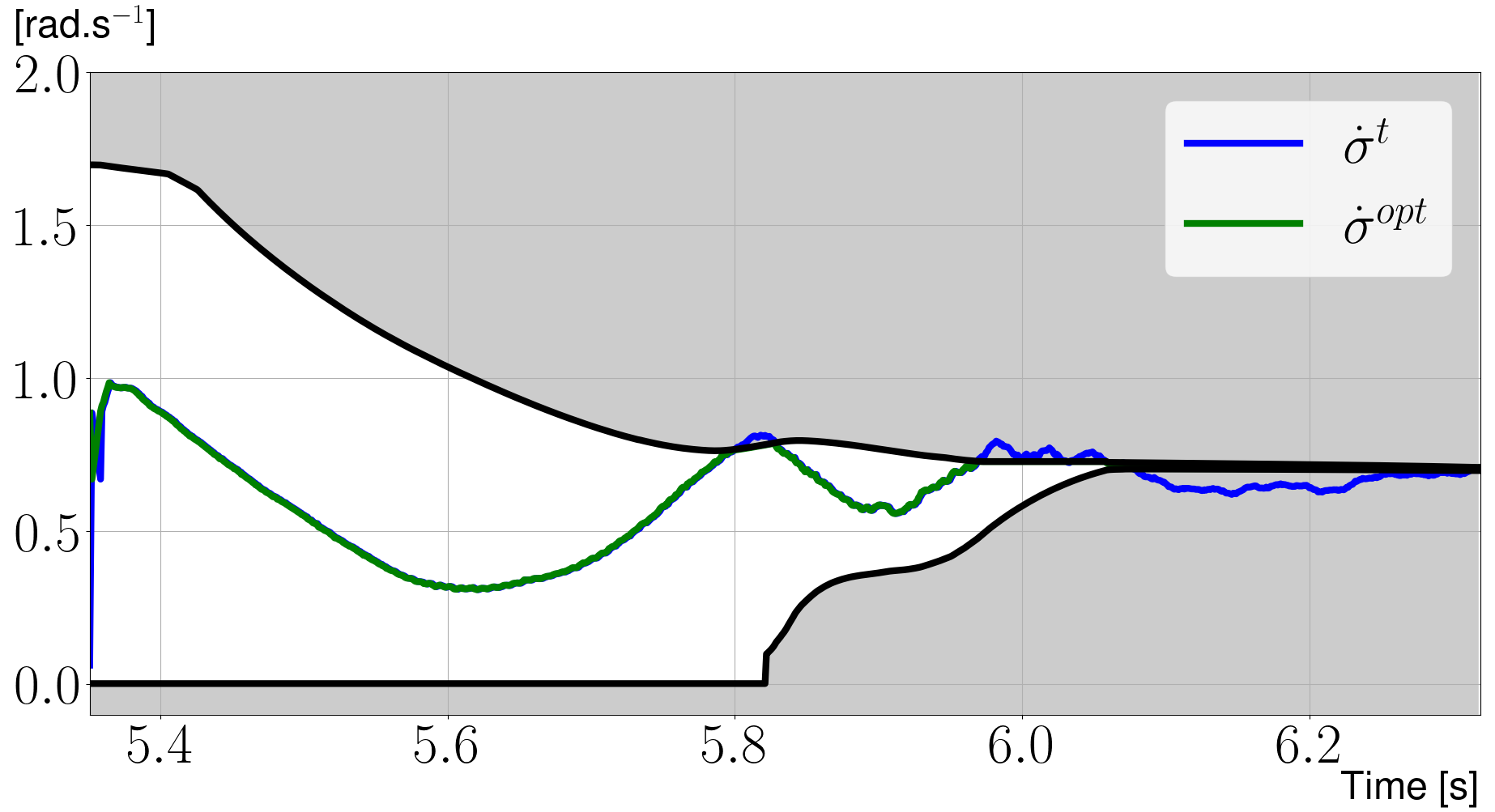}
     \caption{Effective velocity $\optVel$ and \targetVelocity $\targetVel$ over a step with an able-bodied user (step 6 of~\cref{fig:10steps}). Black curves: lower and upper limits of the feasible velocities set.}
    \label{fig:spv}
 \end{figure}

\begin{figure}[tb]
     \begin{subfigure}[b]{0.24\textwidth}
         \centering
         \includegraphics[scale=0.18]{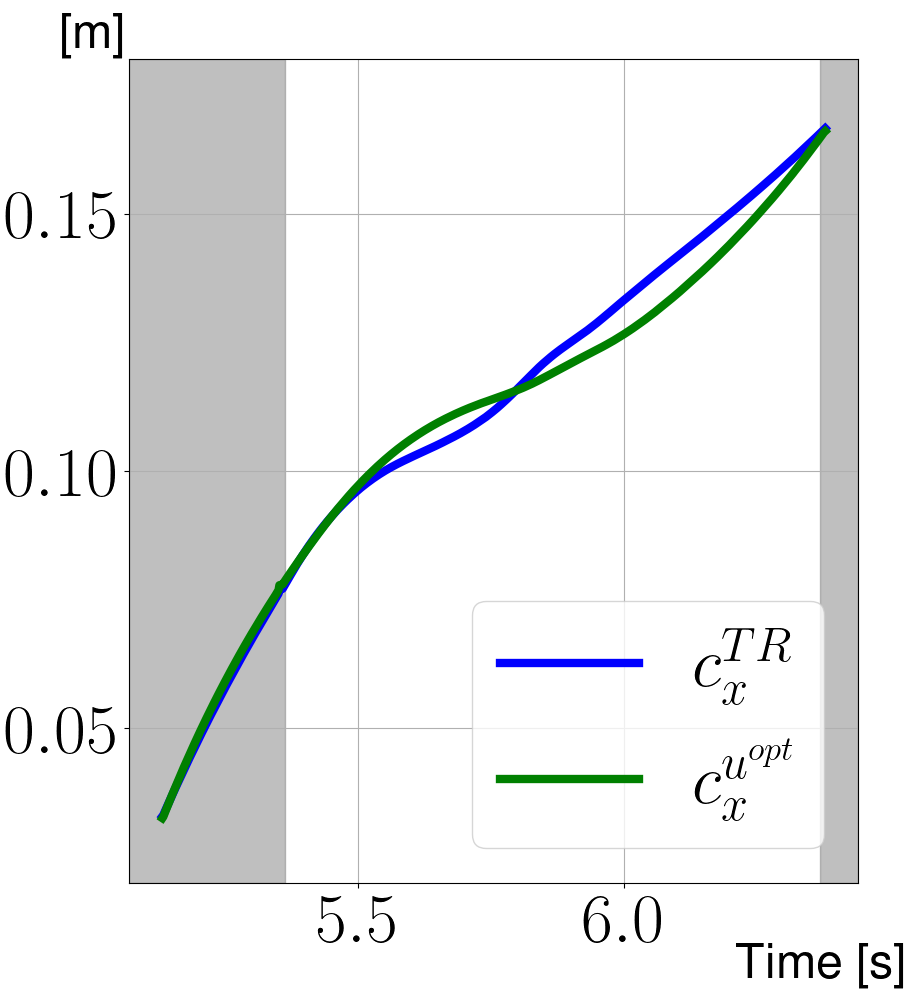}
     \end{subfigure}
     \begin{subfigure}[b]{0.24\textwidth}
         \centering
         \includegraphics[scale=0.18]{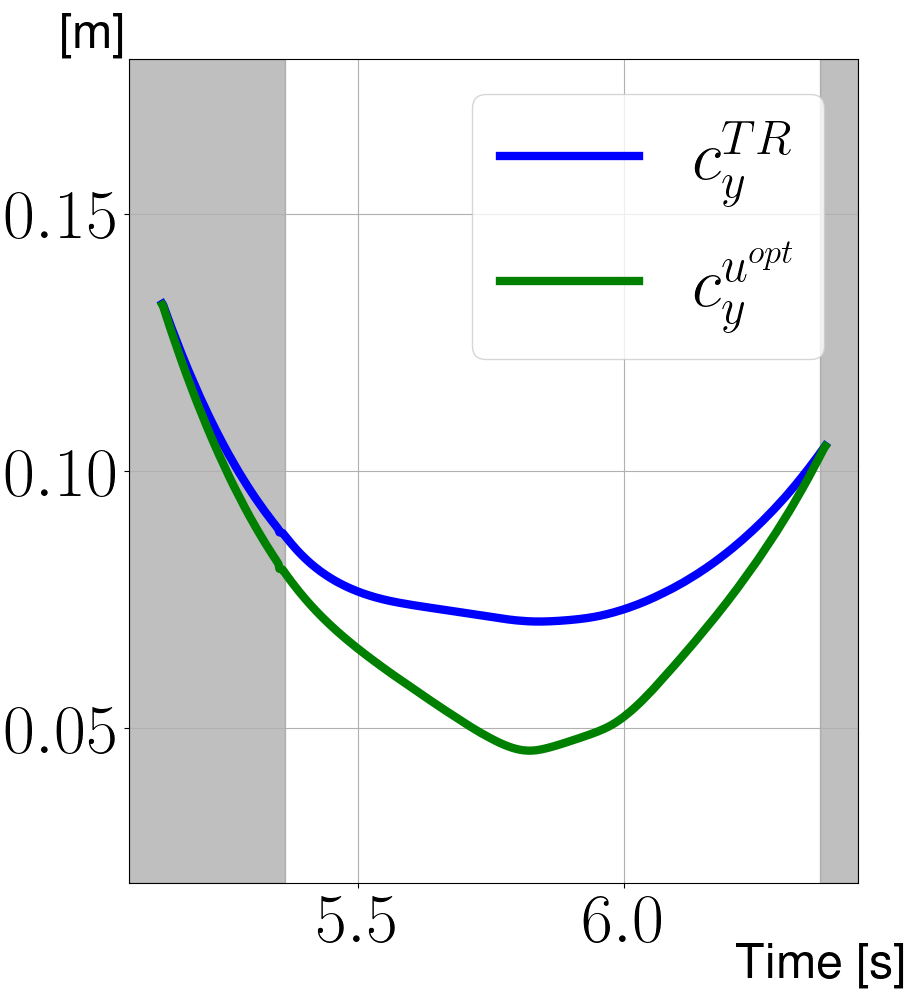}
     \end{subfigure}
      \caption{CoM from TR and OP strategies over a step with an able-bodied user (step 6 of~\cref{fig:10steps}). Grey areas: double support phases. White areas: single support phases. Left: X axis. Right: Y axis.}
     \label{fig:CoM}
  \end{figure}

We report below the results of a 10-steps walking experiment with an able-bodied user using the proposed controller with the OP strategy. A video of the experiments can be viewed at \url{https://youtu.be/_1A-2nLy5ZE}. We first report a single step velocity (on~\cref{fig:spv}) and CoM trajectory (on~\cref{fig:CoM}).
\Cref{fig:spv} reveals how the OP strategy accounts for the \uniLatContConst. In detail, during the first $450ms$ of the single support phase, the OP strategy leaves the \targetVelocity unchanged because the solution of~\cref{PB1} is $\optTime = \targetTime$ ($\optVel$ in green completely overlaps $\targetVel$ in blue). Hence, the patient's schedule is fully respected. During the remaining $500$\,ms, the OP strategy starts filtering the \targetVelocity to preserve the balance of the system, $\sigma^{opt} \neq \targetVel$. Gradually, $\sigma^{opt}$ is constrained around $71\%$ normalized velocity.

The OP strategy also wisely chooses the CoM reference trajectory and satisfies the final state constraint. \Cref{fig:CoM} shows the final constraint is satisfied as the replanned (OP) and time-rescaled (TR) CoM trajectories' endpoints are identical.
The CoM trajectory computed with OP strategy is very different from the one with TR strategy, and, in particular, exhibits a minimum on the Y axis 2.5\,cm closer to the support foot (centered at 0.0\,cm) than the nominal trajectory: the exoskeleton sways its hip toward the support foot to accommodate for the user's low-velocity desire. This is consistent with human behavior. 

\begin{figure}[tb]
    \centering
    \includegraphics[scale=0.18]{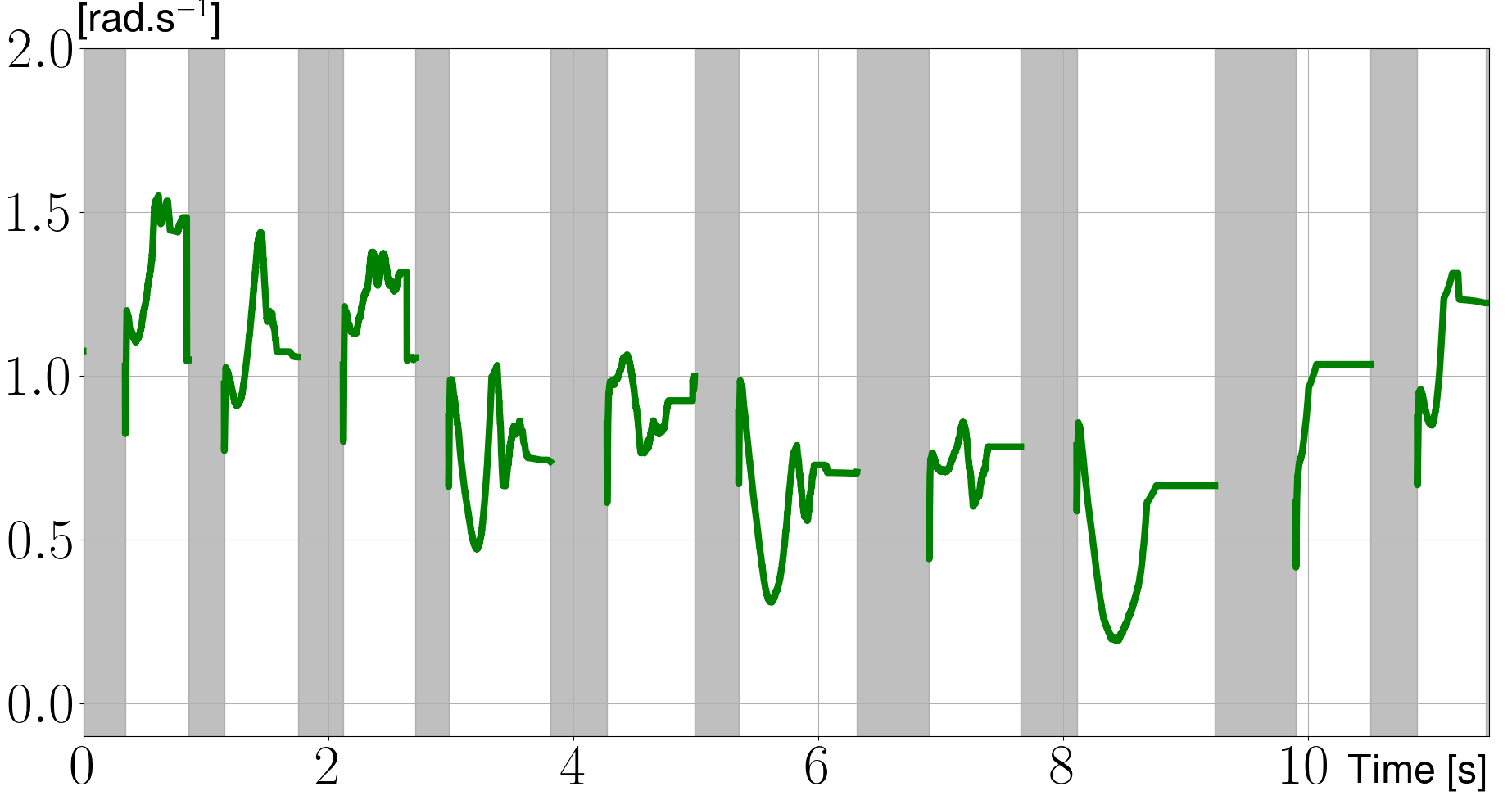}
    \caption{Effective velocity $\optVel$ over a 10-steps experiment with an able-bodied user. Grey areas: double support phases. White areas: single support phases.}
    \label{fig:10steps}
\end{figure}

Finally, \cref{fig:10steps} shows $\optVel$ over the whole experiment, with double support areas in grey. During this experiment, the user varies the level of efforts produced by their legs during the single support phases. Note that~\cref{VG} is only used during these phases while the user's desire is ignored during double support phases\footnote{More precisely, the reference trajectory used during double support is computed once, at the beginning of the step. For this, we use the OP strategy and the mean velocity of the previous step, for sake of continuity}.


\section{Conclusion}
The control architecture proposed in this article extends the functionalities of Atalante and enables rehabilitation tasks for walking impaired patients. Since stability of the walk is handled by the self-balanced exoskeleton, the physiotherapist is relieved from this tedious but critical task. 

The quantitative evaluation on physical health improvement remains to be done. In particular, the evaluation of the controller from a medical perspective will certainly be insightful to guide future developments.

In anticipation, several points in the methodology could be improved further. The new controller is not active during the double support phases of the walk. However, the CoM transfer during double support is a question of interest for walking rehabilitation and could be addressed with further developments of the presented method. It would also be interesting to consider adapting the step length to the patient efforts. This is a very natural extension to be addressed in the online planning strategy. For this, a library of predefined gaits could be used. Finally, the LIP model could be enhanced to address more dynamical gait patterns where the angular momentum variations can not be neglected.

\appendix
\subsection{CoP control using admittance on the support leg}\label{LLCAdmittance}


The admittance scheme is adapted from~\cite{caronStairClimbingStabilization2019} by reorganizing the so-called \emph{Stack-of-Tasks (SoT)}~\cite{escandeHierarchicalQuadraticProgramming2014}. The original $SoT_1$ reads, in decreasing order of priority as follows
\begin{description}
    \item[Level 0:]~ support and swing foot position and velocity tracking;
    \item[Level 1:]~ CoM acceleration tracking;
    \item[Level 2:]~ pelvis roll and pitch tracking;
    \item[Level 3:]~ static standing articular configuration tracking.
\end{description}

We use $SoT_1$ during double support phases, and the $SoT_2$, described below, during single support phases in order to account for the swing leg effect on the CoM acceleration

\begin{description}
    \item[Level 0:]~ support foot and \textit{swing leg articular} position $P(\refSigma)$ and velocity $\refVel T(\refSigma)$ tracking;
    \item[Level 1:]~ CoM acceleration tracking;
    \item[Level 2:]~ pelvis roll and pitch tracking;
    \item[Level 3:]~ static standing articular configuration tracking.
\end{description}

$SoT_1$ and $SoT_2$ are Hierarchical Quadratic Programs. They are solved for the articular target acceleration $\ddot{q}^t$ that best satisfy their objectives. Numerical integration from the previous articular position and velocity targets yields the support leg target position and velocity $({\qsp}^*, \dot{q}^{sp^*})$ tracked using~\eqref{PDsp}. Set-points for Level 2 and 3 of both $SoT_1$ and $SoT_2$ are computed from the nominal gait $\mT$.

We validated this custom $SoT_2$ by performing a performance comparison with $SoT_1$, both experimentally and in simulation, and found no noticeable impact on the overall stability of the walk (their transient responses differ in shape but not in error magnitude).

\printbibliography

\end{document}